\documentclass[letterpaper, 10 pt, conference]{ieeeconf}  

\IEEEoverridecommandlockouts
\usepackage{amsmath} 
\usepackage{amsfonts} 
\usepackage{graphicx}
\usepackage[ruled]{algorithm2e}
\usepackage{multirow}
\usepackage{caption} 
\usepackage{cite}

\DeclareMathOperator*{\argmax}{\arg\max}

\usepackage{graphics} 
\usepackage{graphicx,import} 
\usepackage{epsfig} 
\usepackage{mathptmx} 
\usepackage{amsmath} 
\usepackage{amssymb}  
\usepackage{mathtools}
\DeclarePairedDelimiterX\set[1]\lbrace\rbrace{#1}
\usepackage{hyperref}

\usepackage[percent]{overpic}
\usepackage{xcolor}
\newcommand{\insertYoutubeLink}{\url{https://youtu.be/mv2xw83NyWU}}

\title{\LARGE \bf
Combining Benefits from Trajectory Optimization \\
and Deep Reinforcement Learning
}

\author{Guillaume Bellegarda and Katie Byl%
\thanks{This work was funded in part by NSF NRI award 1526424.}
\thanks{Guillaume Bellegarda and Katie Byl are with the Robotics Laboratory, Department of Electrical and Computer Engineering, University of California at Santa Barbara (UCSB).
        {\tt\small gbellegarda@ucsb.edu, katiebyl@ucsb.edu}}%
}

\begin{document}

\maketitle
\thispagestyle{empty}
\pagestyle{empty}

\begin{abstract}

Recent breakthroughs both in reinforcement learning and trajectory optimization have made significant advances towards real world robotic system deployment. Reinforcement learning (RL) can be applied to many problems without needing any modeling or intuition about the system, at the cost of high sample complexity and the inability to prove any metrics about the learned policies. Trajectory optimization (TO) on the other hand allows for stability and robustness analyses on generated motions and trajectories, but is only as good as the often over-simplified derived model, and may have prohibitively expensive computation times for real-time control. This paper seeks to combine the benefits from these two areas while mitigating their drawbacks by (1) decreasing RL sample complexity by using existing knowledge of the problem with optimal control, and (2) providing an upper bound estimate on the time-to-arrival of the combined learned-optimized policy, allowing online policy deployment at any point in the training process by using the TO as a worst-case scenario action. 
This method is evaluated for a car model, with applicability to any mobile robotic system. A video showing policy execution comparisons can be found at \insertYoutubeLink.
\end{abstract}


\section{Introduction}

Deep reinforcement learning (DRL) methods have shown recent success on continuous control tasks in robotics systems in simulation~\cite{googleEmergence,ppo,lillicrap15}.
Such methods are applied using no prior knowledge of the systems, leading to problematic sample complexity and thus long training times. Unfortunately, little can be said about the stability or robustness of these resulting control policies, even if more traditional model-based optimal control solutions exist for these same systems.

Additionally, DRL has been almost exclusively applied in simulation, where a failed trial has no repercussions. In the real world a failure can have catastrophic consequences, including damaging the robot or causing injury to humans in the area. Some recent works have successfully learned a policy for a real robot~\cite{drl_robotic_manipulation}~\cite{learning_to_walk_drl}, or transferred policies learned in simulation to the real system~\cite{sim2real}~\cite{anymal_sim2real}. Of particular note in~\cite{anymal_sim2real} is that the learned policies outperform the authors' previous model-based methods with respect to both energy-efficiency and speed. However, instead of training from scratch, it would seem intuitive to use the model-based methods as an initial starting point for DRL, with the reasoning that at any given moment, our learned policy should do \emph{at worst as well as} existing control solutions.

\begin{figure}[thpb]	
\centering
	\includegraphics[width = 3.4 in]{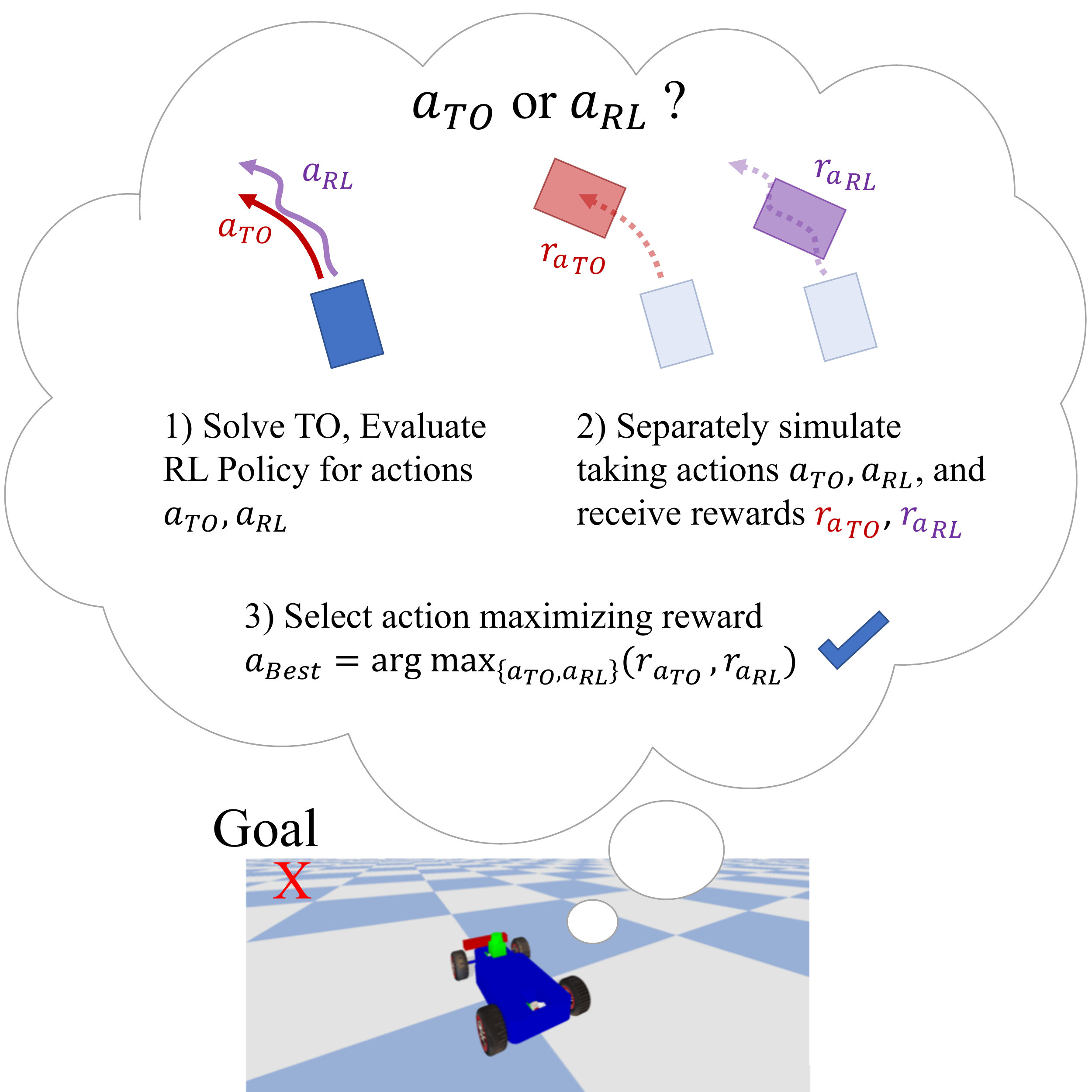}\\
\caption{Agent picking between actions from trajectory optimization or from reinforcement learning, simulating taking each respectively, and then selecting the one leading to greater reward to execute in the real world.} 
\label{intro_graphic}
\end{figure}

One might be tempted to perform imitation learning updates on trajectories taken from running a model-based optimal control policy, for example using DAgger~\cite{DAgger}. However, due to often mismatching dynamics between the simplified system on which this model-based control policy is based off of, and the real physical system, this may lead to overfitting a suboptimal policy. There is also the additional concern of deviating too much from the expert trajectories into regions of the state space not previously visited, in which the policy learned only from expert data may perform poorly.

Instead, we propose interweaving optimal control samples during the policy rollouts of model-free DRL methods in the following manner: at each timestep we can evaluate our policy network to get action $a_{RL}$, as well as query our trajectory optimization to get action $a_{TO}$. 
We then simulate the execution of each of these actions individually, and select the one which gave the larger reward as our true action to use in the real world.
Such a scheme, shown in Figure~\ref{intro_graphic}, should ensure that at worst the agent will always do as well as the model-based optimal control policy, and can only do better.
At the beginning of training, we expect this approach to almost exclusively pick $a_{TO}$; due to the network weights being randomly initialized, it is very unlikely to consistently outperform a model-based method. However as training progresses, and from added policy exploration/exploitation, the number of selected on-policy samples will increase.

Related work on using trajectory optimization to help learn or guide a control policy include~\cite{Mordatch14combiningthe},~\cite{Levine2013variational}, and~\cite{Levine2013gps}. These works differ from the proposed method in this paper as they focus more on incorporating offline demonstrations or trajectories into training to guide the policy search, whereas in this work the trajectory optimization is run online at each timestep and compared with the current policy action, ensuring a worst case scenario.

A related work combining prior knowledge of the system with learning is~\cite{learning_with_training_wheels}, where the policy chooses between actions computed with a simple PID controller and from evaluating the current actor network. Although the authors observe this controller helps achieve faster and more stable learning performance, it is not optimal and much can still be improved in terms of sample complexity. Additionally, there are no guarantees on the policy at any given time, and no minimum time to goal estimates or worst case scenarios.

Another related work that combines learning with MPC is POLO~\cite{polo}. POLO seeks to improve MPC by learning a global value function, of which it has only a local estimate when initialized. As a result, it cannot be run online and provides no policy guarantees, as it cannot achieve a desired result without first learning the global value function.
This paper in contrast seeks to improve on policy learning by using a trajectory optimization framework to guide the learning process, and provides a worst-case scenario action that can be run online.

As a model-based optimal control policy, in this work we take direct inspiration from the work done in~\cite{Posa13, posa2016optimization, Posa14, Feng15}, where full body control for underactuated systems is achieved via trajectory optimization and stabilization under constraints. \cite{posa2016optimization} in particular introduces an algorithm (DIRCON) that extends the direct collocation method, incorporating manifold constraints to produce nominal trajectories with third-order integration accuracy.

The rest of this paper is organized as follows: Section~\ref{background} provides background details on reinforcement learning, imitation learning as behavioral cloning, and robot dynamics in the context of a car model. Section~\ref{sec:TO} describes the trajectory optimization framework used to calculate optimal trajectories for the car, and our algorithm combining this trajectory optimization with deep reinforcement learning (in this case PPO) is presented in Section~\ref{sec:coto-ppo}. Section~\ref{sec:result} shows results on the benefits of using our algorithm, and a brief conclusion is given in Section~\ref{sec:conclusion}.

\section{Preliminaries}
\label{background}

\subsection{Reinforcement Learning}

The reinforcement learning framework, which is described thoroughly in~\cite{sutton_rl_book} and elsewhere, typically consists of an agent interacting with an environment modeled as a Markov Decision Process (MDP). An MDP is given by a 4-tuple $(S,A,T,R)$, where $\emph{S}$ is the set of states, $\emph{A}$ is the set of actions available to the agent, $T: S \times A \times S \rightarrow \mathbb{R}$ is the transition function, where $T(s,a,s')$ gives the probability of being in state $s$, taking action $a$, and ending up in state $s'$, and  $R: S \times A \times S \rightarrow \mathbb{R}$ is the reward function, where $R(s,a,s')$ gives the expected reward for being in state $s$, taking action $a$, and ending up in state $s'$.
The goal of an agent is thus  to interact with the environment by selecting actions that will maximize future rewards.

In this paper, the states consist of a subset of a robot's positions and velocities, the actions are motor torques or positions, the transition function is modeled by a physics engine~\cite{pybullet}, and the reward is a potential-based function to minimize distance to a target goal. 

\subsection{Proximal Policy Optimization}
\label{sec:ppo}

Although we expect to see benefits from combining trajectory optimization with any deep reinforcement learning algorithm,
for this paper we use the current state-of-the-art, Proximal Policy Optimization (PPO)~\cite{ppo}. In particular, PPO has achieved breakthrough results for continuous control robotics tasks by optimizing the following surrogate objective with clipped probability ratio: 
\begin{align}
\label{ppo_obj}
L^{CLIP}(\theta) = \hat{\mathbb{E}}_t [\min(r_t(\theta)\hat{A}_t, \text{clip}(r_t(\theta),1-\epsilon,1+\epsilon)\hat{A}_t]
\end{align}
where $\hat{A}_t$ is an estimator of the advantage function at time step $t$ as in~\cite{Schulman15}, and $r_t(\theta)$ denotes the probability ratio
\begin{align}
\label{ppo_prob_ratio}
r_t(\theta) = \frac 
			{{\pi}_{\theta}(a_t | s_t)} 
            {{\pi}_{\theta_{old}}(a_t | s_t)}
\end{align}
where $\pi_\theta$ is a stochastic policy, and $\theta_{old}$ is the vector of policy parameters before the update.
This objective seeks to penalize too large of a policy update, which means penalizing deviations of $r_t(\theta)$ from 1.

\subsection{Learning from Demonstration}

In this work we use the classical behavioral cloning (BC) approach to imitation learning where we seek to minimize the error between an expert action and the maximum likelihood estimate action from the current policy: 
\begin{align}
    L^{BC}(\theta) = \frac{1}{N} \sum_{i=1}^{N} (\ a_i^* - \argmax_{a_i} \pi_\theta(a_i|s_i)\ )^2 
\end{align}
for expert demonstration state-action pairs $\{s_i, a_i^*\}^N_{i=1}$, where $\argmax_{a_i} \pi_\theta(a_i|s_i)$ is the maximum likelihood estimate action $a_i$ for state $s_i$ using policy $\pi_\theta$.

\subsection{Robot Dynamics}

The equations of motion for a robotic system can be written as:
\begin{align} 
D(q)\ddot{q} + C(q,\dot{q})\dot{q} + G(q) + A(q)^T\lambda = B(q)u + F\label{eq:manip}
\end{align}
\noindent where $q$ are the generalized coordinates, $D(q)$ is the inertial matrix, $C(q,\dot{q})\dot{q}$ denotes centrifugal and Coriolis forces, $G(q)$ captures potentials (gravity), $A(q)^T\lambda$ are constraint forces (where $\lambda$ are unknown multipliers a priori), $B(q)$ maps control inputs $u$ into generalized forces, and $F$ contains non-conservative forces such as friction. 

\begin{figure}[thpb]
      \centering
      \includegraphics[width=2.5in]{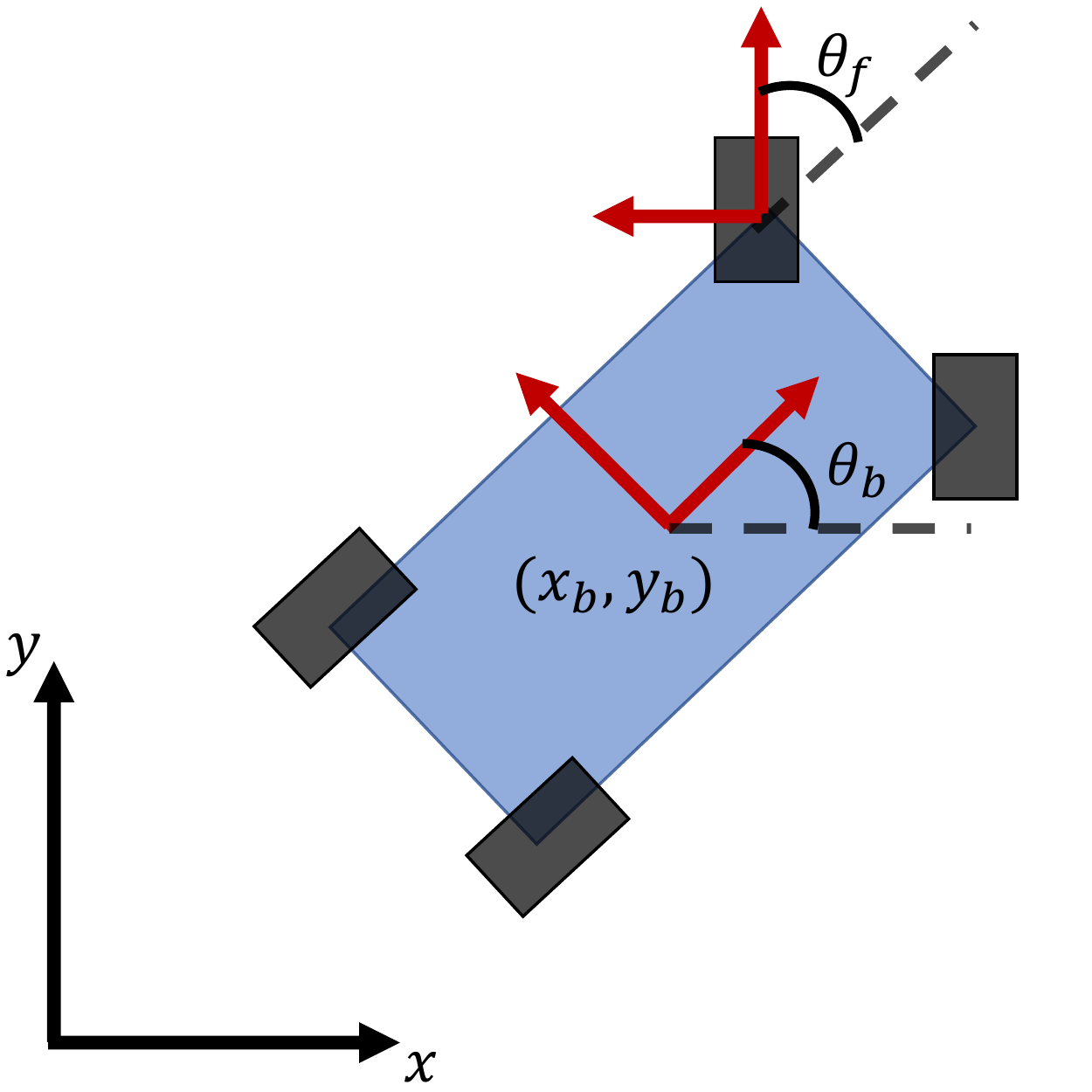} \\
      \caption{Car model used for trajectory optimization. }
      \label{fig:car_gc}
\end{figure}

In this work, we specifically consider a simple car model, shown in Figure~\ref{fig:car_gc}, with $ q = [x_b, y_b, \theta_b, \theta_f]^T $, where $(x_b,y_b)$ are the center of mass coordinates in the world frame, $\theta_b$ is the yaw of the body with respect to the global x-axis, and $\theta_f$ is the steering angle of the front wheels.

For general wheeled mobile robots, $A(q)^T\lambda$ typically contains constraints ensuring no slip (free rolling in the direction the wheel is pointing) and no skid (no velocity along the wheel's rotation axis perpendicular to the free rolling direction), which come from writing these constraints in Pfaffian form $A(q)\dot{q}=0$. $\lambda$ can be explicitly solved for by differentiating $A(q)\dot{q}=0$ and substituting in $\ddot{q}$ from Equation~\ref{eq:manip}.

\subsubsection{Remark} Because this constraint is in place, the trajectory optimization will not find solutions where skidding is a viable option allowing for greater reward (such as skidding into a parking space instead of parallel parking, or an aggressive turning maneuver to more quickly change directions). 
The alternative to this constraint would be to add friction approximations such as in~\cite{trinkle_stewart}, for which the optimization must then solve a Linear Complementarity Problem at each contact point. As this can be a very expensive computation, we avoid this consideration and instead entrust the DRL algorithm to use the trajectory optimization as a guide towards learning a better policy, in which slip may or may not be optimal.

\section{Trajectory Optimization}
\label{sec:TO}

This section provides details for formulating the locomotion problem for a robotic system as a trajectory optimization. At a high level, the full nonlinear system is discretized, and direct collocation along with backward Euler integration is used to generate motion as in~\cite{Posa13}~\cite{cdc19_to}. More precisely, the problem is formulated as:
\begin{align}
\mbox{find}\quad &q,\dot{q}, u \\
\mbox{subject to}\quad & \text{minimize cost } J  \nonumber \\
& \text{State Constraints } \nonumber\\
&\phi(q,\dot{q},u)=0 \\
&\psi(q,\dot{q},u) \geq 0 \\
& \text{Dynamics Constraints } \nonumber\\
&\begin{aligned}
    D(q)\ddot{q} &+ C(q,\dot{q})\dot{q} + G(q) \\
    &+A(q)^T\lambda =  B(q)u + F
\end{aligned}
\end{align}

\noindent where each of the above constraints are detailed below, along with cost function considerations.

\subsection{Objectives}
The cost function $J$ is defined as the weighted squared error between the goal coordinates $(x_g,y_g, \theta_g)$ and the body coordinates $(x_N,y_N, \theta_N)$, where $N$ is the number of sample points for the trajectory:
\begin{align}
J = \alpha(x_g - x_N)^2 + \beta(y_g - y_N)^2 + \gamma(\theta_g - \theta_N)^2
\end{align}
where weights $\alpha, \beta, \gamma$ can vary based on the desired task, i.e. if final body orientation is important.
\subsection{State constraints}
The initial states $q_0$ and $\dot{q}_0$ are constrained exactly based on the robot's current state. For the rest of the $N$ time points, $q$ and $\dot{q}$ are bounded by joint position and velocity limits. The input torques $u$ are also bounded explicitly by the physical constraints of the robot, as well as implicitly by $\dot{q}$ ranges.  
\subsection{Dynamics constraints}
At each time step $k$, with $h=\Delta t$ the time step interval, the dynamics are constrained:
\begin{align}
q_{k+1} = q_k + h\dot{q}_{k+1}  \\
\dot{q}_{k+1} = \dot{q}_k + h\ddot{q}_{k+1}  
\end{align}
with
\begin{align}
\ddot{q}_{k+1} = D_{k+1}^{-1}&(B_{k+1}u_{k+1} + F_{k+1} - C_{k+1}\dot{q}_{k+1} \nonumber \\ &-G_{k+1}-A_{k+1}^T\lambda_{k+1}) 
\end{align}

\noindent where we write $D(q_{k+1})$ as $D_{k+1}$, and similar for other terms.

\section{Cooperative Trajectory Optimization and Deep Reinforcement Learning}
\label{sec:coto-ppo}

In this section we detail our algorithm, Cooperative Trajectory Optimization and PPO (CoTO-PPO), shown in Algorithm 1. The main idea is that for each new observation $s_t$ at time step $t$, the current PPO actor network is queried for action $a_{RL}$, and a trajectory optimization is solved for action $a_{TO}$. Each of these actions is simulated individually to get rewards $r_{a_{RL}k+1}$ and $r_{a_{TO}k+1}$. The action that produces the larger simulated reward is the one that is selected as the true best action and used in the real world (or to step the actual simulation). Necessary transition information is then appended to either the PPO dataset $D_{PPO}$ or Supervised Learning (SL) dataset $D_{SL}$, depending on which action was selected. After $T$ time steps corresponding to the current policy/trajectory optimization roll out, the actor-critic PPO networks are updated by optimizing $L^{CLIP}(\theta)$ on dataset $D_{PPO}$, and the actor network is additionally updated with supervised learning by optimizing $L^{BC}(\theta)$ on dataset $D_{SL}$.

\begin{algorithm}
 Initialize function approximation parameters $\theta$\\
 Initialize PPO and Supervised Learning datasets $D_{PPO}, D_{SL}$\\
 \For{training epoch=1,2,...}{
  \For{timestep=1,2,...T}{
    Solve trajectory optimization for action $a_{TO}$ \\
    Evaluate PPO policy network for $a_{RL} \sim \pi_\theta(a_t,s_t)$\\
    Simulate taking each action separately and select action maximizing next step reward: 
    \begin{align}
        a_{Best} = \argmax_{(a_{TO},a_{RL})} (r_{a_{RL}k+1},r_{a_{TO}k+1})  \nonumber
    \end{align}  \\
    Step environment with $a_{Best}$: $s_{t+1} \sim p(s_{t+1}|s_t,a_{Best})$\\
    \uIf{$a_{Best} ==  a_{RL}$}{
    $D_k \sim $ partial trajectory, transition information\\
    $D_{PPO} = D_{PPO} \cup D_k$\\
    }
    \ElseIf{$a_{Best} ==  a_{TO}$}{
    $D_{SL} = D_{SL} \cup \{(s_t, a_{TO}) \}$
    }
  }
  \For{K epochs on $D_{PPO}$ }{
   normal PPO updates by optimizing $L^{CLIP}(\theta)$
   }
   \For{K epochs on $D_{SL}$}{
   supervised learning LfD updates by optimizing $L^{BC}(\theta)$
   }
 }
 \caption{Cooperative Trajectory Optimization and PPO (CoTO-PPO)}
\end{algorithm}


\section{Results}
\label{sec:result}

\subsection{Implementation Details}

We use a combination of OpenAi Gym~\cite{openaigym} to represent the MDP and PyBullet~\cite{pybullet} as the physics engine for training and simulation purposes.

\begin{figure*}[thpb]
      \centering
      \includegraphics[width=1.45in]{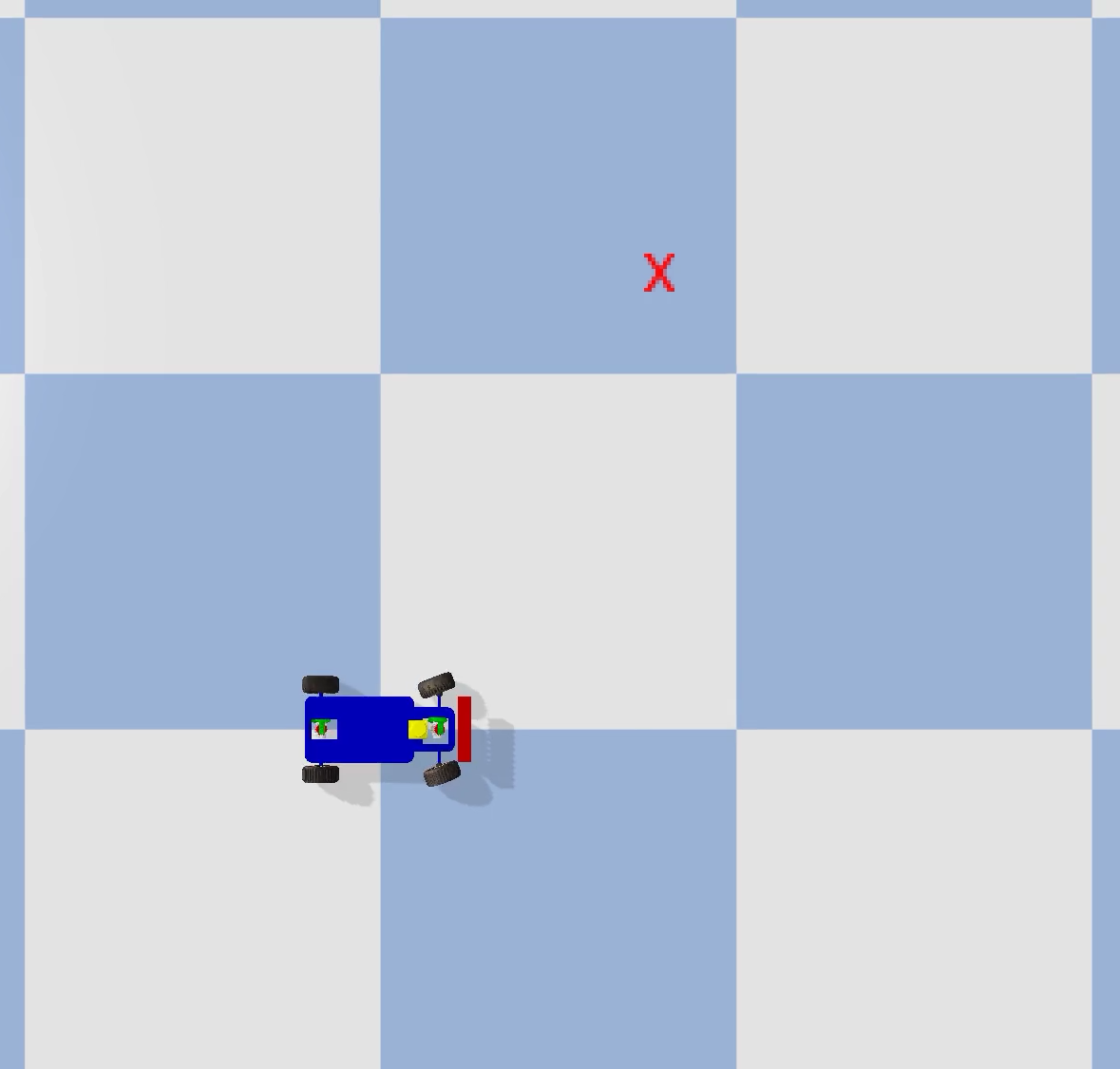}~~\includegraphics[width=1.45in]{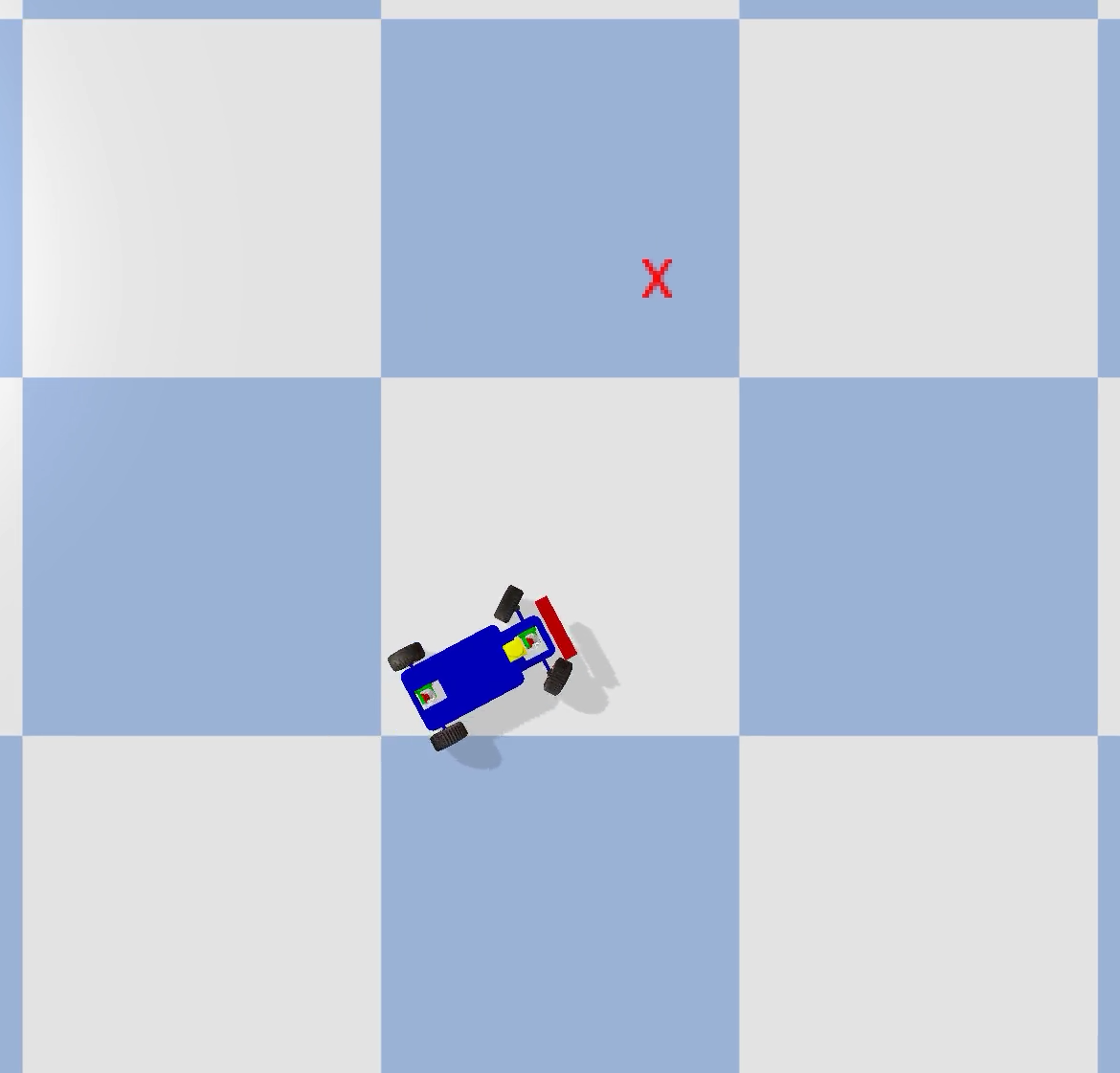}~~\includegraphics[width=1.45in]{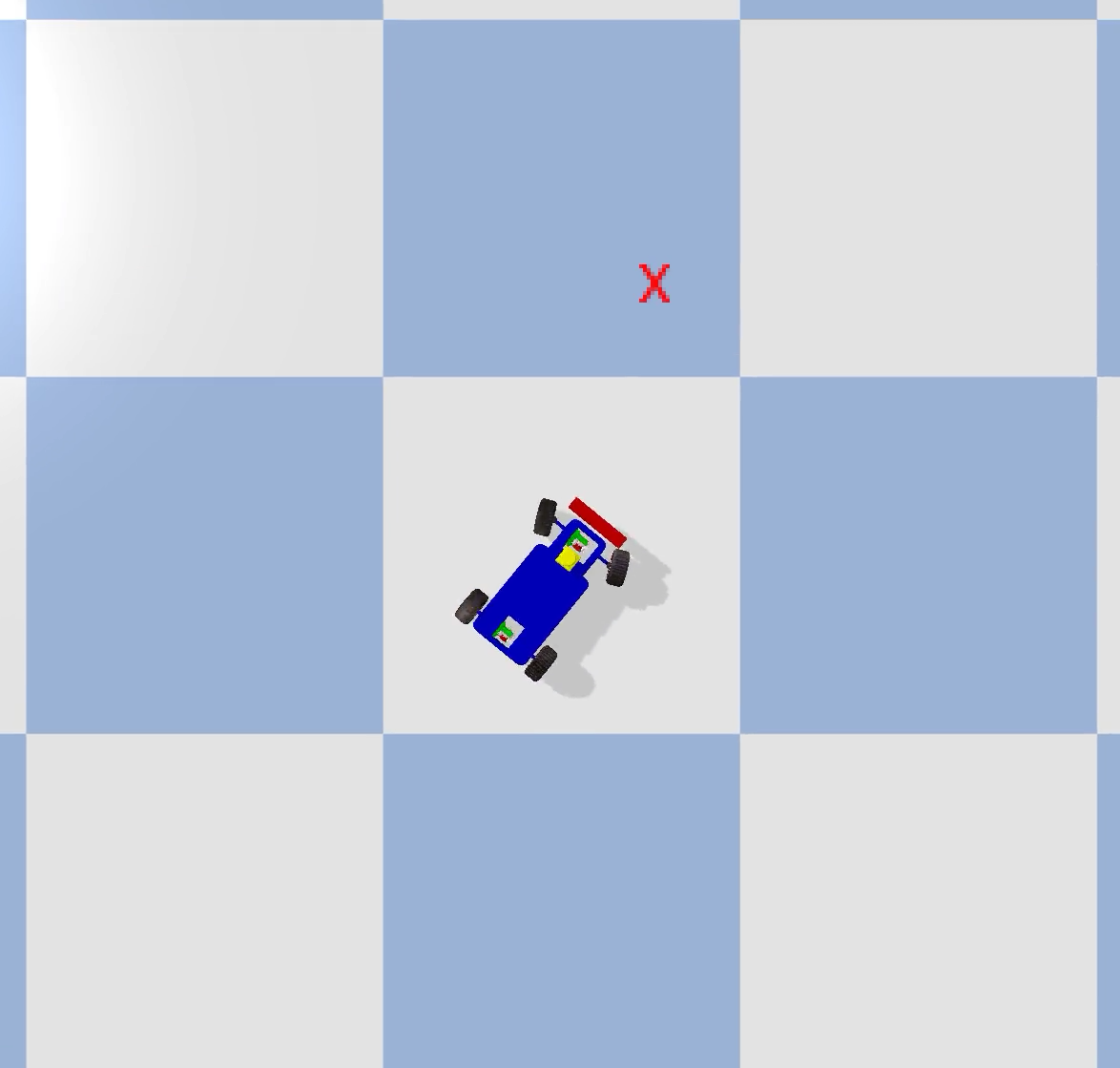}~~\includegraphics[width=1.45in]{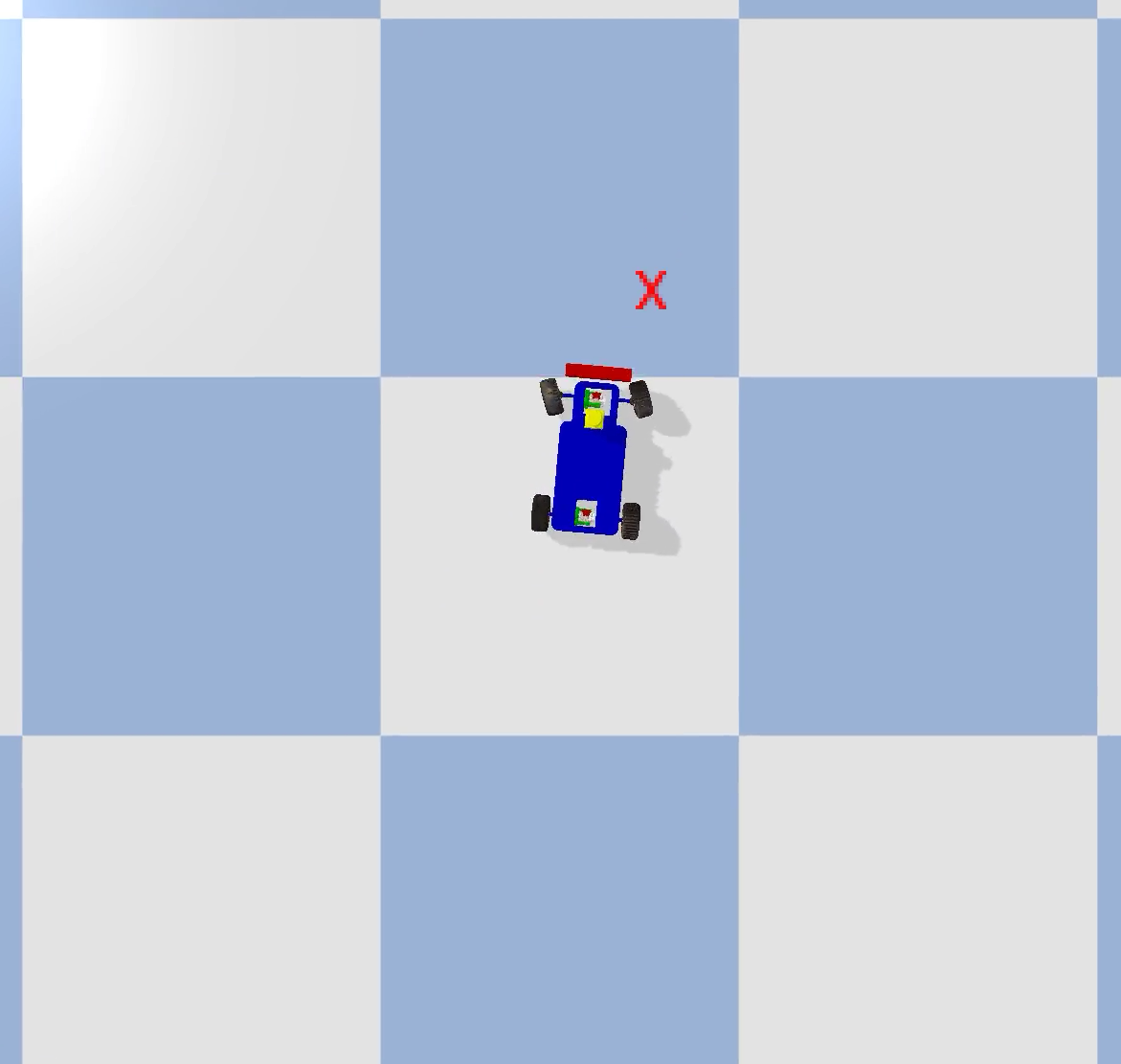} \\
      \caption{ Environment snapshots of Bullet racecar with desired goal denoted by the red `X'.}
      \label{fig:snapshots}
\end{figure*}

We additionally use the OpenAI Baselines~\cite{baselines} implementation of PPO (which optimizes $L^{CLIP}(\theta)$ as discussed in Sec. ~\ref{sec:ppo}) with the default hyperparameters, but with the Beta distribution to select continuous actions as suggested in~\cite{beta_distribution_drl} to avoid the bias introduced with limited control ranges when using the standard Gaussian distribution. The Beta distribution parameters $\alpha$ and $\beta$ are TensorFlow variables and are therefore updated during each SGD minibatch, so the action variance will decrease as the policy converges.
The Gym environment we use is similar to the standard HumanoidFlagRun environments, but the humanoid is replaced with the Bullet MIT racecar. Example snapshots from the environment are shown in Figure~\ref{fig:snapshots}. The goal destination/flag is updated only when the car center of mass lies within 0.2 [m] of the current goal location, and then placed randomly in a 10 by 10 [m] grid.
The agent has 10 seconds to maximize rewards each trial, which will typically consist of reaching several goal locations consecutively. 

Our neural network architecture is the default Multi-Layer Perceptron, consisting of 2 fully connected hidden layers of 64 neurons each, with tanh activation. The policy and value networks each have this same structure. 

The trajectory optimization is implemented in Python with CasADi~\cite{casadi}, using IPOPT~\cite{ipopt} to solve the NLP. Due to the imposed torque and velocity limits as well as the nonholonomic constraints added to the dynamics, the trajectory optimization will find solutions that are suboptimal to the true best policy. We will see in the following subsection that even this suboptimal trajectory optimization has a large benefit when combined with policies either learned from scratch or with our method, both during training and testing.  

The observation space is:
\begin{align}
    [\ ||(x_g,y_g) - (x_b,y_b)||,\ \theta_g - \theta_b,\ \theta_f,\ \dot{x}_b,\ \dot{y}_b,\ \dot{\theta}_b,\ \dot{\theta}_f\ ] 
\end{align}
which is the distance from the center of mass to the goal location, the angle from the current body heading to the goal, the steering angle of the front wheels, the body velocity in the global $x$ and $y$ directions, the body yaw rate, and the yaw rate of the steering wheels.

The action space is $[\ v,\ \theta_f\ ]$, which is a desired body velocity to be set with velocity control mapped to a differential drive, and desired steering angle to be set with position control.

We consider potential-based shaping reward functions of the form: 
\begin{equation} 
\label{eq:potential_fctn}
F(s,a,s') = \gamma \Phi(s') - \Phi(s)
\end{equation}
to guarantee consistency with the optimal policy, as proved by Ng et. al in~\cite{Ng_policy_invariance}. The real valued function $\Phi : S \rightarrow \mathbb{R}$ seeks to minimize the distance to a target goal $(x_g,y_g)$:
\begin{equation}
\label{eq:min_dist}
\Phi(s) = - \sqrt{(x_b-x_g)^2 + (y_b-y_g)^2}
\end{equation}

This reward scheme gives dense rewards at each time step, towards ensuring the optimal policy is learned, rewarding incremental motion in the direction of the current goal. Having dense rewards is important in this framework as we are choosing between actions based on the simulated instantaneous reward, which would likely be 0 at most time steps under sparse reward scenarios.

\subsection{Experiments}

We seek to compare and evaluate the following methods:
\begin{enumerate}
    \item \emph{pure PPO}
    \item \emph{pure trajectory optimization}
    \item \emph{CoTO-PPO} - (our method)
    \item \emph{CoTO-PPO, policy only} - how well does the policy learned from CoTO-PPO perform on its own, without the fail-safe action of the trajectory optimization?
    \item \emph{CoTO-(pure PPO)} - how well does combining trajectory optimization with an entirely separately trained agent with PPO perform?
\end{enumerate}
The reader is encouraged to watch the accompanying video\footnote{\insertYoutubeLink} for simulations of the discussed policies.

\begin{figure}[!thpb]
      \centering
      \includegraphics[width=3.4in]{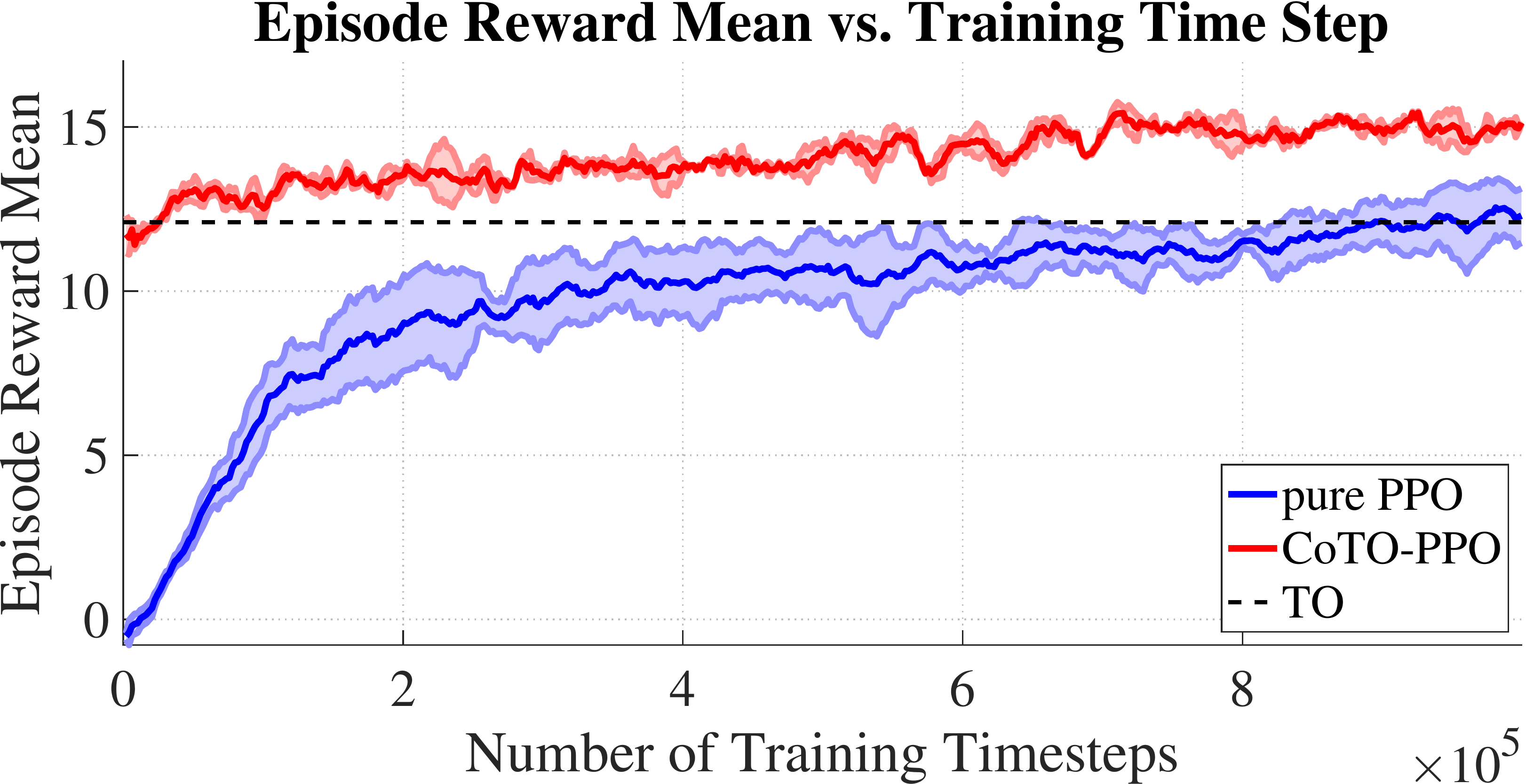} \\
      \caption{Episode reward mean for pure PPO, and cooperative trajectory optimization and PPO (CoTO-PPO). The episode reward mean from using only the trajectory optimization is plotted as a dashed line. }
      \label{fig:eprewmean}
\end{figure}

Figure~\ref{fig:eprewmean} shows the episode reward mean vs. number of training time steps for running pure PPO as well as CoTO-PPO on the CarFlagRun environment. The episode reward mean indicates how well the agent was able to continue to progress in the direction of the goal location(s) during each trial.
Due to doing at worst as well as the trajectory optimization, CoTO-PPO begins with very high reward mean, and only improves from there as the networks are updated with both DRL and supervised learning updates. Pure PPO on the other hand is forced to learn from scratch, and even after training for 1 million time steps, is only able to do as well as the trajectory optimization combined with essentially uniformly distributed random noise of the uninitialized policy from CoTO-PPO.

\begin{figure}[thpb]
      \centering
      \includegraphics[width=3.4in]{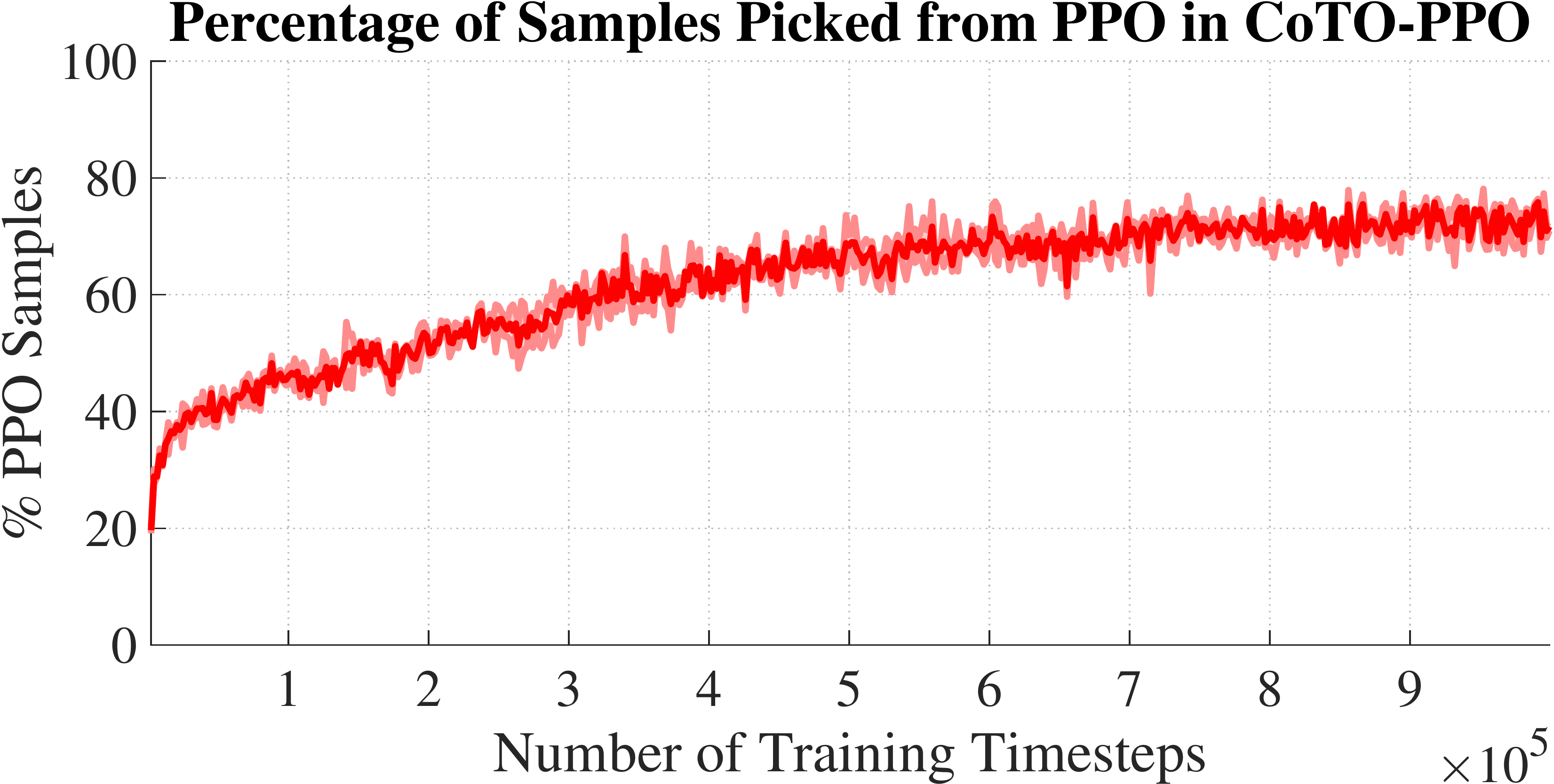} \\
      \caption{ At the beginning of training, about 20\% of the random actions selected by PPO produce larger rewards than the optimal actions found with the trajectory optimization. Over time the policy is guided towards better regions and roughly 75\% of the samples selected from PPO result in maximal reward. As a fail safe, the TO action is taken to keep our bounds on time to goal.}
      \label{fig:rl_samples}
\end{figure}

Figure~\ref{fig:rl_samples} shows the percentage of samples picked by the policy network of PPO in CoTO-PPO that outperform the actions from the trajectory optimization over training. As expected, when the policies are randomly initialized, few samples from PPO will outperform even a suboptimal trajectory optimization. Eventually as training progresses, the percentage of maximal reward samples picked with PPO converges to around 75\% of the time. This shows there is still a benefit to using the trajectory optimization as a worst case scenario, as it is still being picked 25\% of the time after 1 million training time steps.

\begin{table*}[thpb]
\centering
\begin{tabular}{ |c||c|c|c|c|  }
\hline 
 & \multicolumn{2}{|c|}{Reward Mean} & \multicolumn{2}{|c|}{Percent PPO Actions}\\
\hline
Algorithm & Stochastic & Deterministic & Stochastic & Deterministic\\[.2ex]
\hline
pure PPO  & 12.9 & 13.8 & - & - \\
Trajectory Optimization & -  & 12.1 & - & - \\
CoTO-PPO  & 15.1 & \bf{15.7} & 75 & \bf{81} \\
CoTO-PPO, policy only & 14.1 & 14.7 & - & -\\
CoTO-(pure PPO) & 14.5 & 14.5 & 44 & 57\\
 \hline
\end{tabular} \\
\caption{Episode reward mean and percent of samples chosen with PPO for different algorithms across 100 trials, using either stochastic or deterministic (maximum likelihood) actions from the output distributions of the policy network. The percent of PPO actions are only listed for algorithms which choose between both DRL and TO actions, and the trajectory optimization action is always evaluated deterministically.}
\label{table:tab1}
\end{table*}

Table~\ref{table:tab1} details the reward mean and percentage of PPO actions picked (if relevant) with various algorithms and scenarios across 100 trials, after training for 1 million time steps. We do trials of evaluating each policy by sampling from the output Beta distributions stochastically, as well as deterministically evaluating the distributions with the maximum likelihood estimate.
We see that combining the trajectory optimization with PPO significantly increases the mean reward, with our method CoTO-PPO having the best performance. If we evaluate only the policy trained from CoTO-PPO, it is still a significant improvement over the policy trained from pure PPO alone. 
We also evaluate the effect of combining the policy trained with pure PPO with the trajectory optimization, labeled CoTO-(pure PPO), which makes it clear that pure PPO has learned a suboptimal policy, as the combination with the TO leads to a larger reward mean.

In this latter case, we also track what percent of the time CoTO-(pure PPO) picks the TO action vs. the action selected from the policy network of pure PPO. Despite training for 1 million time steps, our algorithm finds that the trajectory optimization performs better than the policy trained from scratch roughly half of the time. In comparison, the policy trained from our method CoTO-PPO is picked over 75\% of the time, despite far fewer on-policy samples (due to using the TO samples for supervised updates).

\subsection{Maximum Instantaneous Reward Discussion}
A first look at the algorithm may seem to imply that it is greedy, rather than optimal, as the agent selects the action leading to the maximum instantaneous reward, rather than a function of expected returns. We experimented with simulating taking multiple actions from the TO and from RL over varying horizons, but found significantly worse performance with this method. One plausible explanation for this result is that when first initialized, the PPO actor network is essentially taking random actions. As the horizon $h$ increases on which we simulate taking $h$ actions from TO and RL separately, the expected returns of taking a series of random actions regresses toward 0 under our reward scheme, and thus the probability that RL will outperform the TO tends to 0. Said another way, it becomes increasingly unlikely to have multiple ``lucky actions'' from RL in a row during exploration as the horizon $h$ increases. Since the agent will correspondingly almost always choose the actions from the suboptimal TO, the policy network will almost always be updated with supervised learning updates and will correspondingly converge, approximately, to this same suboptimal policy. 

On the other hand, by using the maximum instantaneous transition reward from a horizon of 1, there is a much higher probability of sampling a ``good'' action from an uninitialized random policy. This allows for more efficient exploration of the environment than by overwhelmingly following the (suboptimal) trajectory optimization actions, leading to a better overall policy (such as more aggressive throttle values, steering angles during turns, etc.), while still ensuring a reasonable worst case scenario action from the more dynamically conservative TO solution, for cases in which we sample a worse-performing action with RL.
The short horizon also avoids overfitting to the suboptimal trajectory optimization expert.


\section{Conclusion}
\label{sec:conclusion}

In this work we have shown the benefits of combining trajectory optimization and deep reinforcement learning methods into one training process. Using these two methods cooperatively allows for online use of our algorithm at any point in the training process, knowing that the worst case scenario will be as good as a model-based trajectory optimization.
This additionally leads to much greater sample efficiency, and avoids unnecessary exploration of randomly initialized policies, towards avoiding local optima. 

Even if the trajectory optimization is suboptimal due to mismatching dynamics or overly conservative constraints, there is a clear advantage to incorporating prior knowledge of the system to speed up and guide learning. We also observe that trained policies, whether exclusively learned with deep reinforcement learning or from our combined method, are likely to converge to local optima and cannot exhaustively span all observation states, showing the benefit of model-based methods as a proven fail-safe option. The need to be able to put bounds on learned policies and guarantee some sort of behavior is clear, and this work presents preliminary steps in this direction. 

The method detailed in this paper can be readily applied to any robotic system, and should be an effective way to reduce sampling complexity, accelerate training, guide the policy search, deploy policies online at any point in the training process, and give an upper bound estimate on time-to-goal through the trajectory optimization.

\bibliographystyle{IEEEtran}
\bibliography{main.bbl}

\end{document}